\documentclass[
twocolumn,
pre,
floatfix,
superscriptaddress,
]{revtex4-1}

\usepackage[T1]{fontenc}
\usepackage[latin9]{inputenc}
\usepackage{mathrsfs}
\usepackage{amsmath}
\usepackage{amssymb}
\usepackage{graphicx}
\usepackage{braket}
\usepackage{color}
\usepackage[svgnames]{xcolor}
\usepackage{comment}
\usepackage{natbib}
\usepackage{dsfont}
\usepackage{mdframed}

\usepackage{url}
\usepackage{breakurl}
\usepackage[breaklinks]{hyperref}

\makeatletter

\usepackage{listings}

\usepackage{hyperref}

\hypersetup{
  colorlinks=true,
  linkcolor=blue,
  citecolor=ForestGreen,
  urlcolor=DarkOrchid
}

\AtBeginDocument{
  
}

\newcommand{\orcidicon}[1]{\href{https://orcid.org/#1}{\includegraphics[height=\fontcharht\font`\B]{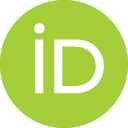}}}

\makeatother

\usepackage{babel}

\begin{document}

\title{Combinatorial Optimization with Physics-Inspired Graph Neural Networks}

\author{Martin~J.~A.~Schuetz \orcidicon{0000-0001-5948-6859}}
\email{maschuet@amazon.com}
\affiliation{Amazon Quantum Solutions Lab, Seattle, Washington 98170, USA}
\affiliation{AWS Intelligent and Advanced Compute Technologies, 
Professional Services, Seattle, Washington 98170, USA}
\affiliation{AWS Center for Quantum Computing, Pasadena, CA 91125, USA}

\author{J.~Kyle~Brubaker \orcidicon{0000-0002-6439-5270}}
\email{johbruba@amazon.com}
\affiliation{AWS Intelligent and Advanced Compute Technologies, 
Professional Services, Seattle, Washington 98170, USA}

\author{Helmut G.~Katzgraber \orcidicon{0000-0003-3341-9943}}
\email{katzgrab@amazon.com}
\affiliation{Amazon Quantum Solutions Lab, Seattle, Washington 98170, USA}
\affiliation{AWS Intelligent and Advanced Compute Technologies, 
Professional Services, Seattle, Washington 98170, USA}
\affiliation{AWS Center for Quantum Computing, Pasadena, CA 91125, USA}

\date{\today}

\begin{abstract}

Combinatorial optimization problems are pervasive across science and
industry. Modern deep learning tools are poised to solve these problems
at unprecedented scales, but a unifying framework that incorporates
insights from statistical physics is still outstanding.  Here we
demonstrate how graph neural networks can be used to solve combinatorial
optimization problems. Our approach is broadly applicable to canonical
NP-hard problems in the form of quadratic unconstrained binary
optimization problems, such as maximum cut, minimum vertex cover,
maximum independent set, as well as Ising spin glasses and higher-order
generalizations thereof in the form of polynomial unconstrained binary
optimization problems. We apply a relaxation strategy to the problem
Hamiltonian to generate a differentiable loss function with which we
train the graph neural network and apply a simple projection to integer
variables once the unsupervised training process has completed.  We
showcase our approach with numerical results for the canonical maximum
cut and maximum independent set problems.  We find that the graph neural
network optimizer performs on par or outperforms existing solvers, with
the ability to scale beyond the state of the art to problems with
millions of variables.

\end{abstract}

\date{\today}

\maketitle

\section{Introduction}
\label{Introduction}

Optimization is ubiquitous across science and industry. Specifically,
the field of combinatorial optimization---the search for the minimum of
an objective function within a finite but often large set of candidate
solutions---is one of the most important areas in the field of
optimization, with practical (yet notoriously challenging) applications
found in virtually every industry, including both the private and public
sectors, as well as in areas such as transportation and logistics,
telecommunications, and finance \citep{glover:18,
kochenberger:14, anthony:17, papadimitriou:98, korte:12}.  While
efficient specialized algorithms exist for specific use cases, most
optimization problems remain intractable, especially in real-world
applications where problems are more structured and thus require
additional steps to make them amenable to traditional optimization
techniques. Despite remarkable advances in both algorithms and computing
power, significant yet generic improvements have remained elusive,
generating an increased interest in new optimization approaches that are
broadly applicable and radically different from traditional operations
research tools.

In the broader physics community, the advent of quantum annealing
devices such as the D-Wave Systems Inc.~quantum annealers
\citep{johnson:11, bunyk:14, katzgraber:18a, hauke:20} has spawned a
renewed interest in the development of heuristic approaches to solve
discrete optimization problems.  On the one hand, recent advances in
quantum science and technology have inspired the development of novel
classical algorithms, sometimes dubbed nature-inspired or
physics-inspired algorithms (e.g., simulated quantum annealing
\cite{kadowaki:98, farhi:01} running on conventional CMOS hardware) that
have raised the bar for emerging quantum annealing hardware; see, for
example, Refs.~\citep{mandra:16b, mandra:18, barzegar:18, hibat:21x}.
On the other hand, in parallel to these algorithmic developments,
substantial progress has been made in recent years on the development of
programmable special-purpose devices based on alternative technologies,
such as the coherent Ising machine based on optical parametric
oscillators \citep{wang:13b, hamerly:18x}, digital MemComputing machines
based on self-organizing logic gates \citep{diventra:18, traversa:15},
and the ASIC-based Fujitsu Digital Annealer \citep{matsubara:17,
tsukamoto:17, aramon:19}.  Some of these approaches face severe
scalability limitations.  For example, in the coherent Ising machine
there is a trade off between precision and the number of variables and
the Fujitsu Digital Annealer --- baked into an ASIC --- can currently
handle at most 8192 variables. Thus, it is of much interest to find new
alternate approaches to tackle large-scale combinatorial optimization
problems, going far beyond what is currently accessible with quantum and
nature-inspired approaches alike.

\begin{figure}[b]
\includegraphics[width=1.0 \columnwidth]{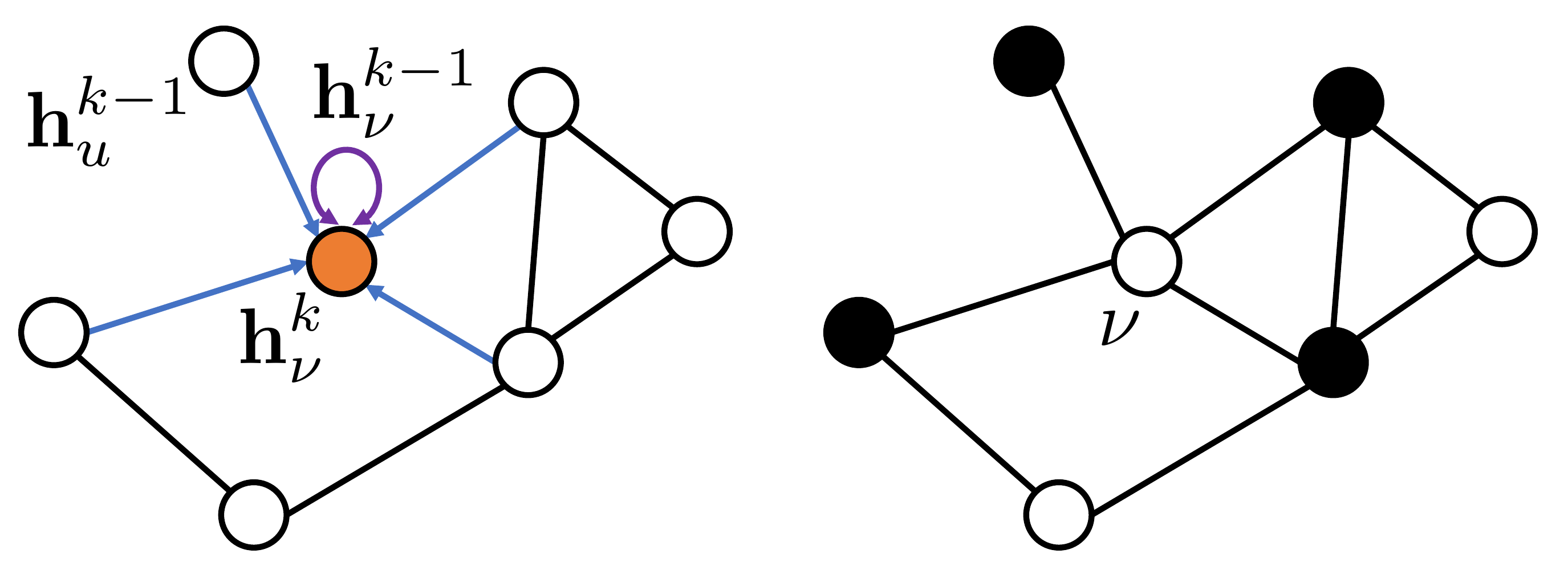}
\caption{
Schematic illustration of the graph neural network approach for
combinatorial optimization presented in this work.  Following a
recursive neighborhood aggregation scheme, the graph neural network is
iteratively trained against a custom loss function that encodes the
specific optimization problem (e.g., maximum cut). At training
completion, we project the final values for the soft node
assignments at the final graph neural network layer back to binary
variables $x_{i}=0,1$, providing the solution bit string
$\mathbf{x}=(x_{1}, x_{2}, \dots)$. Further details are given in the
text. 
\label{fig:scheme}}
\end{figure}

In the deep learning community, graph neural networks (GNNs) have seen a
burst in popularity over the last few years \citep{gori:05,
scarselli:08, micheli:09, duvenaud:15, hamilton:17, xu:19, kipf:17,
wu:19}.  In essence, GNNs are deep neural network architectures
specifically designed for graph structure data, with the ability to
learn effective feature representations of nodes, edges, or even entire
graphs. Prime examples of GNN applications include classification of
users in social networks \citep{perozzi:14, sun:18}, the prediction of
future interactions in recommender systems \citep{ying:18}, and the
prediction of certain properties of molecular graphs \citep{strokach:20,
gaudelet:20}. As a convenient and general framework to model a variety
of real-world complex structural data, GNNs have successfully been
applied to a broad set of problems, including recommender systems in
social media and e-commerce \citep{pal:20, rossi:20}, the detection of
misinformation (fake news) in social media \citep{monti:19}, and various
domains of natural sciences including event classification in particle
physics \citep{choma:18, shlomi:20}, to name a few.  While several
specific implementations of GNNs exist \citep{li:16, velickovic:18,
kipf:17}, at their core typically GNNs iteratively update the features
of the nodes of a graph by aggregating the information from their
neighbors (often referred to as \textit{message passing}
\citep{gilmer:17}) thereby iteratively making local updates to the graph
structure as the training of the network progresses. Because of their
scalability and inherent graph-based design, GNNs present an alternate
platform to build large-scale combinatorial heuristics.

In this work we present a highly-scalable GNN-based solver to
(approximately) solve combinatorial optimization problems with up to
millions of variables.  The approach is schematically depicted in
Fig.~\ref{fig:scheme}, and works as follows: First, we identify the
Hamiltonian (cost function) $H$ that encodes the optimization problem in
terms of binary decision variables $x_{\nu}\in \{0,1\}$ and we associate
this variable with a vertex $\nu \in \mathcal{V}$ for an undirected
graph $\mathcal{G}=(\mathcal{V}, \mathcal{E})$ with vertex set
$\mathcal{V}=\{1, 2, \ldots, n\}$ and the edge set
$\mathcal{E}=\{(i,j):i, j \in \mathcal{V}\}$ capturing interactions
between the decision variables. We then apply a relaxation strategy to
the problem Hamiltonian to generate a differentiable loss function with
which we perform unsupervised training on the node representations of
the GNN. The GNN follows a standard recursive neighborhood aggregation
scheme \citep{gilmer:17, xu:18}, where each node $\nu=1, 2, \ldots, n$
collects information (encoded as feature vectors) of its neighbors to
compute its new feature vector $\mathbf{h}_{\nu}^{k}$ at layer $k=0, 1,
\ldots, K$. After $k$ iterations of aggregation, a node is represented
by its transformed feature vector $\mathbf{h}_{\nu}^{k}$, which captures
the structural information within the node's $k$-hop neighborhood
\citep{xu:19}. For binary classification tasks we typically use
convolutional aggregation steps, followed by the application of a
nonlinear softmax activation function to shrink down the final
embeddings $\mathbf{h}_{\nu}^{K}$ to one-dimensional soft
(probabilistic) node assignments $p_{\nu} = \mathbf{h}_{\nu}^{K} \in
[0,1]$.  Finally, once the unsupervised training process has completed,
we apply a projection heuristic to map these soft assignments $p_{\nu}$
back to integer variables $x_{\nu} \in \{0,1\}$ using, for example,
$x_{\nu}=\mathrm{int}(p_{\nu})$.  We numerically showcase our approach
with results for canonical NP-hard optimization problems such as maximum
cut (MaxCut) and maximum independent set (MIS), showing that our
GNN-based approach can perform on par or even better than existing
well-established solvers, while being broadly applicable to a large
class of optimization problems. Further, the scalability of our approach
opens up the possibility of studying unprecedented problem sizes with
hundreds of millions of nodes when leveraging distributed training in a
mini-batch fashion on a cluster of machines as demonstrated recently in
Ref.~\citep{zheng:20}.

The paper is structured as follows.  In Sec.~\ref{Related Work} we
provide some context for our work, discussing recent developments at the
cross-section between machine learning and combinatorial optimization.
Section \ref{Preliminaries} summarizes the basic concepts underlying our
approach, as well as information on the class of problems that this
approach can solve.  Section \ref{GNN} outlines the implementation of
the proposed GNN-based optimizer, followed by numerical experiments in
Sec.~\ref{Numerics}.  
In Sec.~\ref{Applications} we discuss potential real-world applications in industry.
In Sec.~\ref{Conclusion} we draw conclusions and
give an outlook on future directions of research.

\section{Related Work} 
\label{Related Work}

In this Section we briefly review relevant existing literature, with the
goal to provide a detailed context for our work.  Broadly speaking, our
work makes a \textit{physics-inspired} contribution to the emerging
cross-fertilization between combinatorial optimization and machine
learning, where the development of novel deep learning architectures has
sparked a renewed interest in heuristics for solving NP-hard
combinatorial optimization problems using neural networks, as
extensively reviewed in e.g., Refs.~\citep{kotary:21, cappart:21}.
Leaving alternative, non-graph-based approaches as presented for example
in Ref.~\citep{mills:20} aside, in the following short survey we focus
on graph-based optimization problems---where modern deep learning
architectures such as sequence models, attention mechanisms, and GNNs
provide a natural tool set \citep{kotary:21}---and we primarily
distinguish between approaches based on supervised learning,
reinforcement learning, or unsupervised learning. This categorization
can be refined further with respect to the typical size of a problem
solved by a specific approach and the scope of the solver
(special-purpose vs general-purpose).

\textbf{Supervised Learning.} The majority of neural-network-based
approaches to combinatorial optimization are based on supervised
learning, with the goal to approximate some (typically complex,
non-linear) mapping from an input representation of the problem to the
target solution, based on the minimization of some empirical,
handcrafted loss function. Early work was based on pointer networks
which leverage sequence-to-sequence models to produce permutations over
inputs of variable size, as, for example, relevant for the canonical
traveling salesman problem (TSP) \citep{vinyals:15}. Since then,
numerous studies have fused GNNs with various heuristics and search
procedures to solve \textit{specific} combinatorial optimization
problems, such as quadratic assignment \citep{nowak:17}, graph matching
\citep{bai:18}, graph coloring \citep{lemos:19}, and the TSP
\citep{li:18g, joshi:19}.  As pointed out in Ref.~\citep{karalias:20},
however, the viability and performance of supervised approaches
critically depends on the existence of large, labelled training data
sets with previously optimized hard problem instances, resulting in a
problematic chicken-and-egg scenario, that is further amplified by the
fact that it is hard to efficiently sample unbiased and representative
labeled instances of NP-hard problems \citep{yehuda:20}.

\textbf{Reinforcement Learning.} The critical need for training
labels can be circumvented with Reinforcement Learning (RL) techniques
that aim to learn a policy with the goal of maximizing some expected
reward function. Specifically, optimization problems can typically be
described with a native objective function that can then serve as a
reward function in an RL approach \citep{kotary:21}.  Motivated by the
challenges associated with the need for optimal target solutions, Bello
{\em et al}.~extended the pointer network architecture
\citep{vinyals:15} to an actor-critic RL framework to train an
approximate TSP solver, using a recurrent neural network encoder scheme
and the expected tour length as a reward signal \citep{bello:17}.  Using
a general RL framework based on a graph attention network architecture
\citep{velickovic:18}, significant improvements in accuracy on 
two-dimensional euclidean TSP have subsequently been presented in
Ref.~\citep{kool:19}, getting close to optimal results for problems up
to $100$ nodes.  Moreover, TSP variants with hard constraints have been
analyzed in Ref.~\citep{ma:19}, with the help of a multi-level RL
framework in which each layer of a hierarchy learns a different policy,
and from which actions can then be sampled. Finally, while the majority
of the RL-based approaches have focused on the TSP or variants thereof,
Dai {\em et al.}~proposed a combination of RL and graph embedding to learn
efficient greedy meta-heuristics to incrementally construct a solution,
and showcased their approach with numerical results for Minimum Vertex
Cover, MaxCut, and TSP as test problems, for graphs with up to $\sim
1000$ -- $1200$ nodes \citep{dai:18}.

\textbf{Unsupervised Learning.} Conceptually, our work is most similar
to those that aim to train neural networks in an unsupervised,
end-to-end fashion, without the need for labelled training sets
\citep{karalias:20}.  Specifically, Toenshoff {\em et al}.~have recently
used a recurrent GNN architecture---dubbed RUN-CSP---to solve
optimization problems that can be framed as maximum constraint
satisfaction problems  \citep{toenshoff:19}.  For other types of
problems, such as the maximum independent set problem, the model relies
on empirically-selected hand-crafted loss functions. Using the language
of constraint satisfaction problems, where the system size is expressed
in terms of both the number of variables and the number of constraints,
the authors solve problem instances of Maximum 2-satisfiability,
3-colorability, MaxCut and Maximum Independent Set with up to $5000$
nodes, showing that RUN-CSP can compete with traditional approaches like
greedy heuristics or semi-definite programming. Finally, by either
optimizing a smooth relaxation of the cut objective or applying a policy
gradient, Yao {\em et al}.~trained a GNN to specifically solve the
MaxCut problem, albeit at relatively small system sizes with up to
$500$ nodes \citep{yao:19}, and without any details on runtime.

Here, we present a highly-scalable, physics-inspired framework that uses
deep-learning tools in the form of GNNs to approximate solutions to hard
combinatorial optimization problems with up to millions of variables.
Our GNN optimizer is based on a direct mathematical relation between
prototypical Ising spin Hamiltonians \cite{ising:25}, the Quadratic
Binary Unconstrained Optimization (QUBO) and Polynomial Binary
Unconstrained Optimization (PUBO) formalism and the differentiable loss
function with which we train the GNN, thereby providing one unifying
framework for a broad class of combinatorial optimization problems, and
opening up the powerful toolbox of statistical physics to modern
deep-learning approaches. Fusing concepts from statistical physics with
modern machine learning tooling, we propose a simple, generic, and 
robust solver that does not rely on hand-crafted loss functions.
Specifically, we show that the same GNN optimizer can solve different
QUBO problems, without any need to change the architecture or loss
function, while scaling to problem instances orders of magnitude larger
than what many traditional QUBO solvers can handle
\cite{johnson:11, mandra:16b, matsubara:18, hamerly:19}.

\begin{figure*}
\includegraphics[width=2.0 \columnwidth]{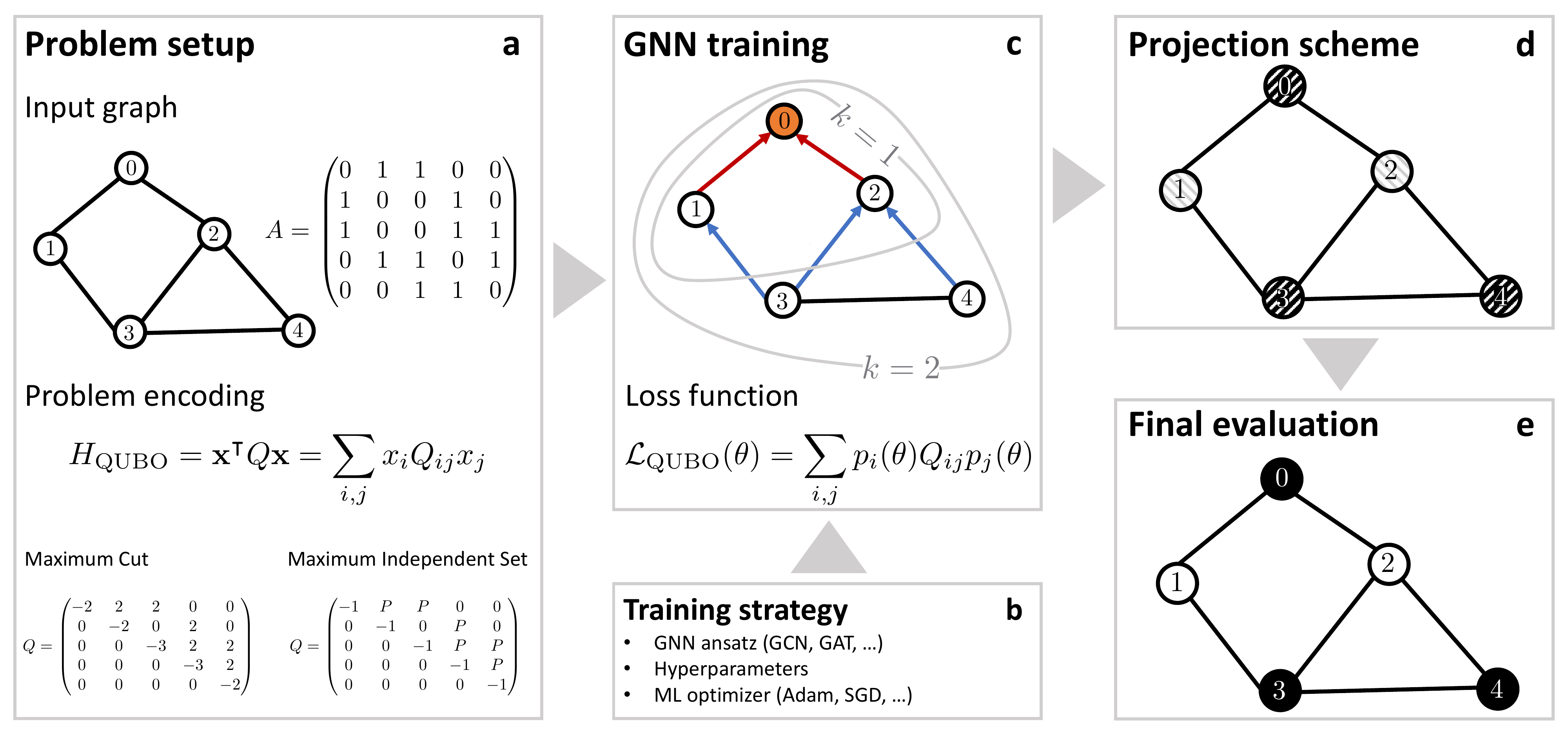}
\caption{
Flow chart illustrating the end-to-end workflow for the proposed
physics-inspired GNN optimizer.  \textbf{(a)}, The problem is specified
by a graph $\mathcal{G}$ with associated adjacency matrix $A$, and a
cost function as described (for example) by the QUBO Hamiltonian
$H_{\mathrm{QUBO}}$.  Within the QUBO framework the cost function is
fully captured by the QUBO matrix $Q$, as illustrated for both MaxCut
and MIS for a sample (undirected) graph with five vertices and six
edges.  \textbf{(b)}, The problem setup is complemented by a training
strategy that specifies the GNN Ansatz, a choice of hyperparameters and
a specific ML optimizer.  \textbf{(c)}, The GNN is iteratively trained
against a custom loss function $\mathcal{L}_{\mathrm{QUBO}}(\theta)$
that encodes a relaxed version of the underlying optimization problem as
specified by the cost function $H_{\mathrm{QUBO}}$.  Typically, a GNN
layer operates by aggregating information within the local one-hop
neighbourhood (as illustrated by the $k=1$ circle for the top node with
label $0$).  By stacking layers one can extend the receptive field of
each node, thereby allowing distant propagation of information (as
illustrated by the $k=2$ circle for the top node with label $0$).
\textbf{(d)-(e)}, The GNN generates soft node assignments which can be
viewed as class probabilities.  Using some projection scheme, we then
project the soft node assignments back to (hard) binary variables $x_{i}
= 0,1$ (as indicated by the binary black/white node coloring), providing the final solution bit string $\mathbf{x}$.
\label{fig:workflow}
}
\end{figure*}

\section{Preliminaries}
\label{Preliminaries}

To set up our notation and terminology we start out with a brief review
of both combinatorial optimization, and graph neural networks.

\textbf{Combinatorial Optimization.} The field of combinatorial
optimization is concerned with settings where a large number of yes/no
decisions must be made and each set of decisions yields a corresponding
objective function value, like a cost or profit value, that is to be
optimized \citep{glover:18}. Canonical combinatorial optimization
problems include, among others, the maximum cut problem (MaxCut), the
maximum independent set problem (MIS), the minimum vertex cover problem,
the maximum clique problem and the set cover problem.  In all cases
exact solutions are not feasible for sufficiently-large systems due to
the exponential growth of the solution space as the number of variables
$n$ increases. Bespoke (approximate) algorithms to solve these problems
can typically be identified, at the cost of limited scope and
generalizability.  Conversely, in recent years the QUBO framework has
resulted in a powerful approach that unifies a rich variety of these
NP-hard combinatorial optimization problems \citep{lucas:14,
kochenberger:14, glover:18, anthony:17}. The cost function for a QUBO
problem can be expressed in compact form with the following Hamiltonian
\begin{equation}
H_\mathrm{QUBO} = \mathbf{x}^{\intercal}Q\mathbf{x} = \sum_{i,j} x_{i}Q_{ij}x_{j}, \label{eq:H-QUBO}
\end{equation}
where $\mathbf{x}=(x_{1}, x_{2}, \dots)$ is a vector of binary decision
variables and the QUBO matrix $Q$ is a square matrix of constant numbers
that encodes the actual problem to solve. Without loss of generality,
the $Q$-matrix can be assumed to be symmetric or in upper triangular
form \citep{glover:18}. We have omitted any irrelevant constant terms,
as well as any linear terms as these can always be absorbed into the
$Q$-matrix because $x_{i}^2=x_{i}$ for binary variables $x_{i} \in
\{0,1\}$.  Problem constraints, as relevant for many real-world
optimization problems, can be accounted for with the help of penalty
terms entering the objective function (rather than being explicitly
imposed), as detailed in Ref.~\citep{glover:18}. The significance of
QUBO problems is further illustrated by the close relation to the famous
Ising model, which is known to provide mathematical formulations for
many NP-complete and NP-hard problems, including all of Karp's 21
NP-complete problems \citep{lucas:14}. As opposed to QUBO problems,
Ising problems are described in terms of binary spin variables
$z_{i}\in\{-1,1\}$, that can be mapped straightforwardly to their
equivalent QUBO form, and vice versa, using $z_{i}=2x_{i}-1$. By
definition, both the QUBO and the Ising models are quadratic, but can be
naturally generalized to higher order PUBO problems, as described by the
$N$-local Hamiltonian
\begin{equation}
H_\mathrm{PUBO} =  \sum_{k=0}^{N}\sum_{\left<i_{1}, i_{2}, \dots, i_{k}\right>} Q_{i_{1}i_{2}\cdots i_{k}} x_{i_{1}} x_{i_{2}} \cdots x_{i_{k}},
\end{equation}
with real-numbered coefficients $Q_{i_{1}i_{2}\cdots i_{k}}$, for some
$N\geq3$, and $\left<i_{1}, i_{2}, \dots, i_{k}\right>$ indicating a
group of $k$ binary variables (or spins in the Ising formulation).
Terms containing a product of $k$ variables, of the form
$Q_{i_{1}i_{2}\cdots i_{k}} x_{i_{1}} x_{i_{2}} \cdots x_{i_{k}}$, are
commonly referred to as $k$-local interactions with
$Q_{i_{1}i_{2}\cdots i_{k}}$ being the coupling constant.  As we
exemplify below for some canonical problems, graph (hypergraph) problems
can be naturally framed as QUBO (PUBO) problems.  To this end, given an
undirected graph $\mathcal{G}=(\mathcal{V}, \mathcal{E})$, we simply
associate a binary variable $x_{i}$ with every vertex $i \in V$, and
then express the (node classification) objective as a QUBO problem,
where the specific assignment $\mathbf{x}$ can be visualized as a
specific two-tone (e.g., light and dark) coloring of the graph
\citep{coloring}; see Fig.~\ref{fig:scheme}.

\textbf{Graph Neural Networks.} On a high level, GNNs are a family of
neural networks capable of learning how to aggregate information in
graphs for the purpose of representation learning.  Typically, a GNN
layer is comprised of three functions \citep{gaudelet:20}: (i) a message
passing function that permits information exchange between nodes over
edges, (ii) an aggregation function that combines the collection of
received messages into a single, fixed-length representation, and (iii)
a (typically nonlinear) update activation function that produces
node-level representations given the previous layer representation and
the aggregated information.  While a single-layer GNN encapsulates a
node's features based on its immediate or one-hop neighborhood, by
stacking multiple layers, the model can propagate each node's features
through intermediate nodes, analogous to the broadening the receptive
field in downstream layers of convolutional neural networks.  Formally,
at layer $k=0$, each node $\nu \in \mathcal {V}$ is represented by some
initial representation $\mathbf{h}_{\nu}^{0} \in \mathbb{R}^{d_{0}}$,
usually derived from the node's label or given input features of
dimensionality $d_0$ \citep{alon:21}.  Following a recursive
neighborhood aggregation scheme, the GNN then iteratively updates each
node's representation, in general described by some parametric function
$f_{\theta}^{k}$, resulting in
\begin{equation}
\mathbf{h}_{\nu}^{k} = f_{\theta}^{k} \left(\mathbf{h}_{\nu}^{k-1}, \{\mathbf{h}_{u}^{k-1} | u \in \mathcal{N}_{\nu}\} \right),  \label{eq:GNN}
\end{equation}
for the layers $k=1, \dots, K$, with $\mathcal{N}_{\nu} = \{ u \in
\mathcal{V} | (u,\nu) \in \mathcal{E} \}$ referring to the local
neighborhood of node $\nu$, i.e., the set of nodes that share edges with
node $\nu$. The total number of layers $K$ is usually determined
empirically as a hyperparameter, as are the intermediate representation
dimensionality $d_k$. Both can be optimized in an outer loop.  While a
growing number of possible implementations for GNN architectures
\citep{wu:19} exists, here we use a graph convolutional network (GCN)
\citep{kipf:17} for which Eq.~\eqref{eq:GNN} reads explicitly as
\begin{equation}
\mathbf{h}_{\nu}^{k} = \sigma \left(\mathbf{W}_{k} \sum_{u \in \mathcal{N}(\nu)} \frac{\mathbf{h}_{u}^{k-1}}{|\mathcal{N}(\nu)|} + \mathbf{B}_{k} \mathbf{h}_{\nu}^{k-1}\right), \label{eq:GCN}
\end{equation}
with $\mathbf{W}_{k}$ and $\mathbf{B}_{k}$ being (shared) trainable
weight matrices, the denominator $|\mathcal{N}(\nu)|$ serving as
normalization factor (with other choices available as well) and
$\sigma(\cdot)$ being some (component-wise) nonlinear activation function
such as sigmoid or ReLU. While GNNs can be used for various prediction
tasks (including node classification, link prediction, community
detection, network similarity, or graph classification), here we focus on
node classification, where usually the last ($K$-th) layer's output is
used to predict a label $y_{\nu}$ for every node $\nu \in \mathcal{V}$.
To this end, we feed the (parametrized) final node embeddings
$\mathbf{z}_{\nu} = \mathbf{h}_{\nu}^{K}(\theta)$ into a
problem-specific loss function and run stochastic gradient descent to
train the weight parameters.

\section{Combinatorial optimization with Graph Neural Networks} 
\label{GNN}

\begin{figure}
\includegraphics[width=1.0 \columnwidth]{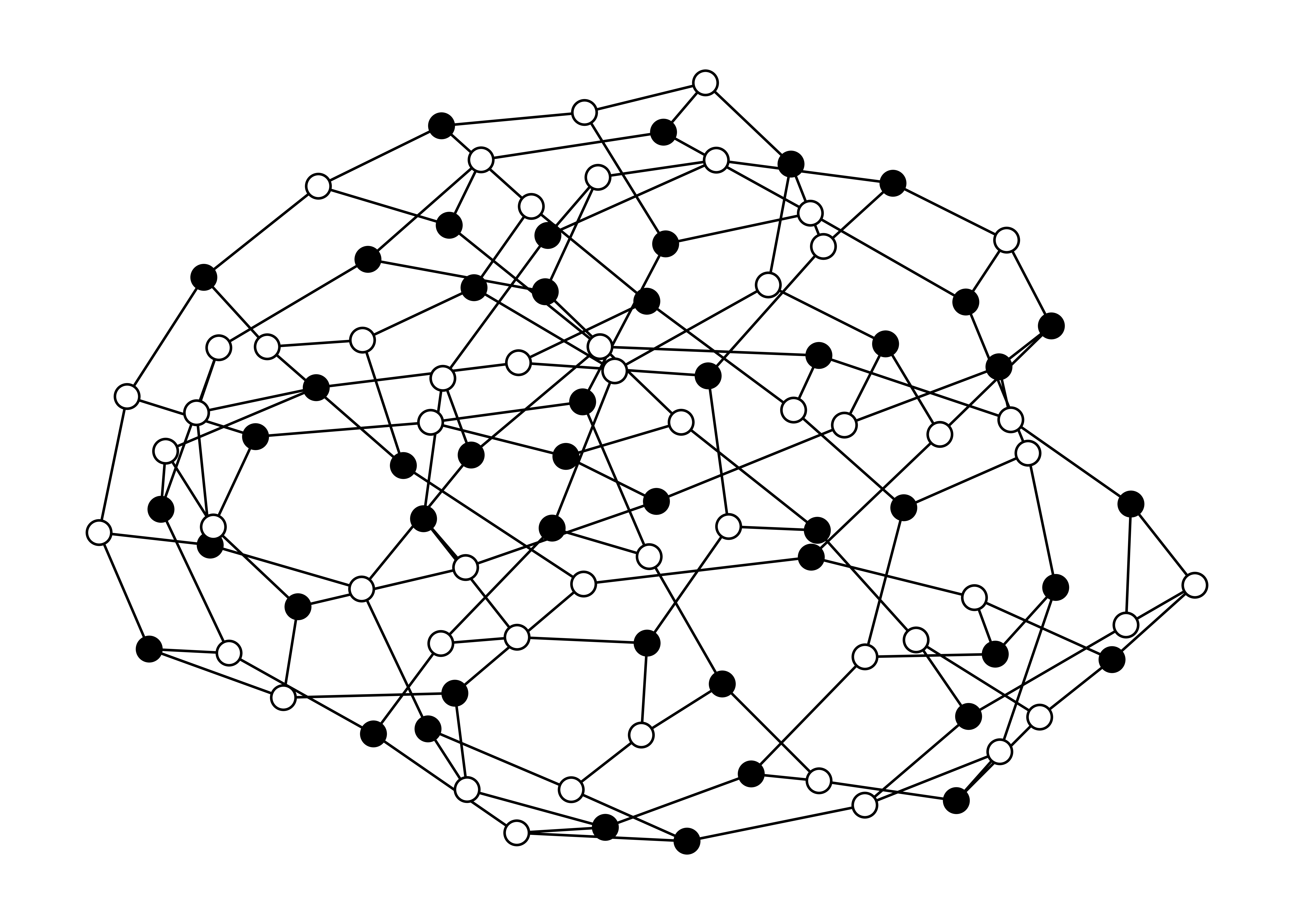}
\caption{
Example solution to MaxCut for a random $3$-regular graph with $n=100$
nodes. After training completion, the GNN provides a binary bit string
$\mathbf{x}$ that assigns one of two possible colors (e.g., black or
white) to each vertex. An edge is said to be \textit{cut} when it
connects two vertices of different colors. For a given graph, the
optimization problem is to assign the colors in a way that as many edges
as possible can be cut at the same time (corresponding to the
antiferromagnetic ground-state of the system).
\label{fig:graph-example-solution}}
\end{figure}

We now detail how to use GNNs to solve combinatorial optimization
problems, as schematically outlined in Fig.~\ref{fig:workflow}. 
To this end, we frame combinatorial optimization problems as
unsupervised node classification tasks, without the need for any
labelled data. Because the nodes do not carry any inherent features, in
our setup the node embeddings $\mathbf{h}_{\nu}^{0}$ are
initialized randomly.  Warm-starting the training process with
pre-training (transfer learning) will be left for future research. The
class of Hamiltonians described above are not differentiable and cannot
be used straightforwardly within the GNN training process. Therefore,
for a given problem Hamiltonian $H$ and graph $\mathcal{G}$, we generate
a differentiable loss function $\mathcal{L}(\theta)$, as required for
standard back-propagation, by promoting the binary decision variables
$x_{i} \in \{0,1\}$ to continuous (parametrized) probability parameters
$p_{i}(\theta)$ with the following (heuristic) \textit{relaxation}
approach
\begin{equation}
x_{i} \longrightarrow p_{i}(\theta) \in [0, 1].
\end{equation}
The soft assignments $p_{i}$ can be viewed as class probabilities. They
are generated by our GNN Ansatz as final node embeddings $p_{i} =
\mathbf{h}_{i}^{K} \in [0,1]$ at layer $K$, after the application of a
non-linear softmax activation function. Then, they are used as input for
the loss function $\mathcal{L}(\theta)$. In particular, for QUBO-type
problems:
\begin{equation}
H_\mathrm{QUBO} \longrightarrow \mathcal{L}_\mathrm{QUBO} (\theta) = \sum_{i,j} p_{i}(\theta)Q_{ij}p_{j}(\theta),  \label{eq:loss-QUBO}
\end{equation} 
which is differentiable with respect to the parameters of the GNN model
$\theta$, and similarly for PUBO problems on hyper-graphs with
higher-order terms of the form $p_{i}p_{j}p_{k}$ etc., thereby
establishing a straightforward, general connection between combinatorial
optimization problems, Ising Hamiltonians and GNNs. For training with
gradient descent, standard ML optimizers such as ADAM can be used.  Once
the (unsupervised) training process has completed, we apply projection
heuristics to map these soft assignments $p_{i}$ back to integer
variables $x_{i}=0,1$, using for example simply
$x_{i}=\mathrm{int}(p_{i})$.  The application of other, more
sophisticated projection schemes will be left for future research.  Note
that any projection heuristics can be applied throughout training after
every epoch, thereby increasing the pool of solution candidates, at no
additional computational cost. With the GNN guiding the search through
the solution space, one can then book keep all solution candidates
identified throughout training and simply pick the best solution found.

Our general GNN approach features several hyperparameters, including the
number of layers $K$, the dimensionality of the embedding vectors
$\mathbf{h}_{i}^{k}$, and the learning rate $\beta$, with details
depending on the specific architecture and optimizer used. These can be
fine-tuned and optimized in an outer loop, using, e.g., standard
techniques such as grid search or more advanced Bayesian optimization
methods.

Our GNN-based approach can be readily implemented with open-source
libraries such as PyTorch Geometric \citep{fey:19} or the Deep Graph
Library \citep{dgl:19}. The core of the corresponding code is displayed
in the supplemental material for a GCN with two layers and a loss
function for any QUBO problem. For illustration, an example solution to
the archetypal MaxCut problem (as implemented with this Ansatz) for a
$3$-regular graph with $n=100$ vertices is shown in
Fig.~\ref{fig:graph-example-solution}. Here, the cut size achieved with
our GNN method amounts to $132$.  Further details are provided below.

\section{Numerical Experiments} \label{Numerics}

We perform numerical experiments using MaxCut and MIS benchmark
problems.  Before providing details on these numerical experiments, we
first describe our GNN model architecture as it is consistent across the
$d$-regular MaxCut and MIS problem instances described below. 
It is certainly
possible that better solutions can be found by fine-tuning the
hyper-parameters for every given problem instance. However, one of 
our goals is to design a robust and scalable solver that is able to
solve a large sample of instances efficiently without the need of
hand-tuning the parameters on an instance-by-instance base.

\textbf{GNN Architecture.} We use a simple two-layer GCN architecture
based on PyTorch GraphConv units. The first convolutional layer is fed
the node embeddings of dimension $d_{0}$ and outputs a representation of
size $d_{1}$. Next, we apply a component-wise, non-linear ReLU
transformation. The second convolutional layer is then fed this
intermediate representation and outputs the output layer of size
$d_{2}$, which is then fed through the component-wise sigmoid
transformation to provide a soft probability $p_{i}\in[0,1]$ for every
node $i \in \mathcal{V}$. We find that the following simple heuristic
for determining the hyper-parameters $d_{0}$ and $d_{1}$ works well: if
the number of nodes is large ($n \geq 10^{5}$), then we set $d_{0} =
\mathrm{int}(\sqrt{n})$, else we set $d_{0} =
\mathrm{int}(\sqrt[3]{n})$, and we take $d_{1} = \mathrm{int}(d_{0}/2)$.
Because we solve for binary classification tasks, we set the final
output dimension as $d_{2} = 1$. However, for multi-color problems this
could be extended to $C>2$ classes by passing the output layer through a
softmax transformation (instead of a sigmoid) and taking the argmax.
Note that as the graph size scales beyond $\sim 10^{5}$ nodes, memory
becomes a concern, and so we further reduce the representations to allow
the GNN to be trained on a single GPU. Distributed training leveraging a
whole cluster of machines will be discussed below in
Sec.~\ref{Conclusion}. With the GNN's output depending on the random
initialization of the hidden feature vectors there is a risk of becoming
stuck in a local optimum where the GNN stops learning. To counter this
issue, one can take multiple shots (i.e., run the GNN training multiple
times for different random seeds and choose the best solution), thereby
boosting the performance at the cost of extended runtime. In our
numerical experiments we limited the number of shots per instance to
five, only re-running the training when an obviously sub-optimal
solution was detected. Finally, we set the learning rate to
$\beta=10^{-4}$ and allow the model to train for up to $\sim 10^{5}$
epochs, with a simple early stopping rule set to an absolute tolerance
of $10^{-4}$ and a patience of $10^{3}$.

\textbf{Maximum Cut.} MaxCut is an NP-hard combinatorial optimization
problem with practical applications in machine scheduling
\citep{alidaee:94}, image recognition \citep{neven:08}, and electronic 
circuit layout design \citep{deza:94}. In the current era of noisy
intermediate-scale quantum devices, with the advent of novel hybrid
quantum-classical algorithms such as the Quantum Approximate
Optimization Algorithm (QAOA) \citep{farhi:14}, the MaxCut problem has
recently attracted considerable attention as a potential use case of
pre-error-corrected quantum devices, see Refs.~\citep{zhou:20,
guerreschi:19, crooks:18, lotshaw:21, patti:21, zhao:20}.  MaxCut is a graph partitioning
problem defined as follows: given a graph with vertex set $\mathcal{V}$
and edge set $\mathcal{E}$, we seek a partition of $V$ into two subsets
with maximum cut, where a cut refers to edges connecting two nodes from 
different vertex sets.  Intuitively, that means we score a point
whenever an edge connects two nodes of different colors. To formulate
MaxCut mathematically, we introduce binary variables satisfying
$x_{i}=1$ if vertex $i$ is in one set and $x_{i}=0$ if it is in the
other set.  It is then easy to verify that the quantity
$x_{i}+x_{j}-2x_{i}x_{j} = 1$ if the edge $(i,j)$ has been cut, and $0$
otherwise.  With the help of the adjacency matrix $A_{ij}$ with
$A_{ij}=0$ if edge $(i,j)$ does not exist and $A_{ij}>0$ if a (possibly
weighted) edge connects node $i$ with $j$, the MaxCut problem is
described by the following quadratic Hamiltonian
\begin{equation}
H_{\mathrm{MaxCut}} = \sum_{i<j} A_{ij}(2x_{i}x_{j}-x_{i} - x_{j}) 
\end{equation}
that falls into the broader class of QUBO problems described by
Eq.~\eqref{eq:H-QUBO}; we provide the explicit $Q$-matrix for
a sample MaxCut problem in Fig.~\ref{fig:workflow}.
Up to an irrelevant constant, the MaxCut problem can equivalently by
described by the compact Ising Hamiltonian $H_{\mathrm{MaxCut}} =
\sum_{i<j}J_{ij}z_{i} z_{j}$ with $J_{ij} = A_{ij}/2$, favoring
antiferromagnetic ordering of the spins for $J_{ij}>0$, as expected
intuitively based on the problem definition. As our figure of merit, we
denote the largest cut found as
$\mathrm{cut}^{\star}=-H_{\mathrm{MaxCut}}(\mathbf{x}^{\star})$, with
$\mathbf{x}^{\star}$ referring to the corresponding bit string.

\begin{figure*}
\includegraphics[width=2.0 \columnwidth]{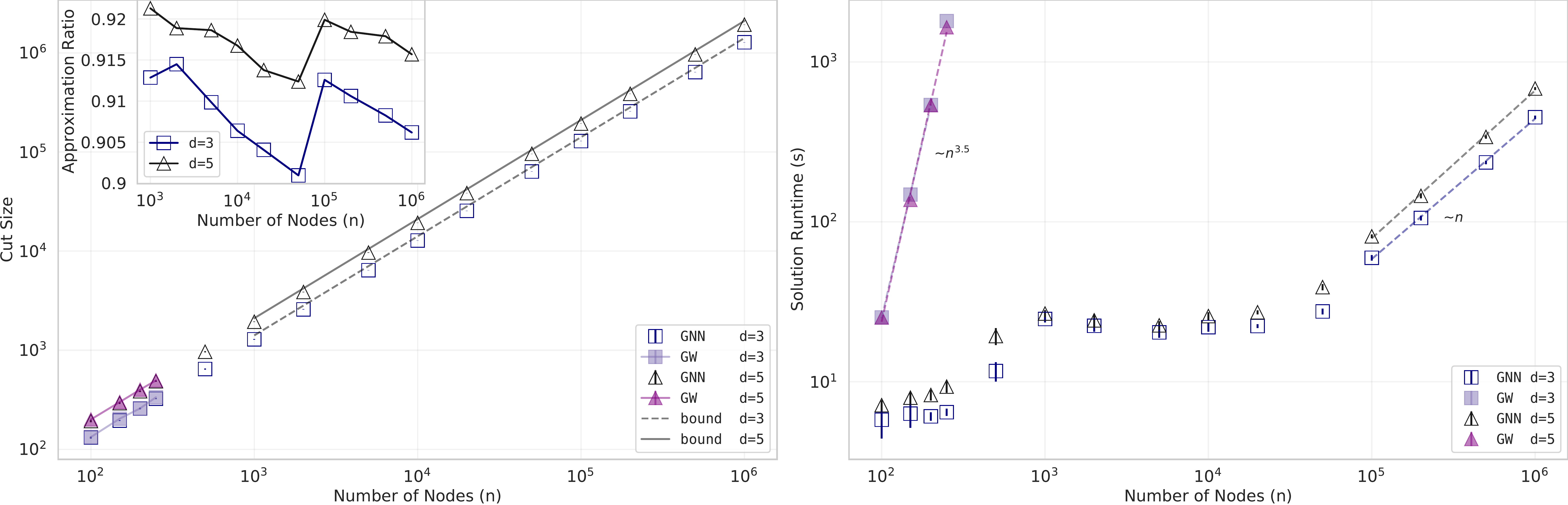}
\caption{
Numerical results for MaxCut. \textbf{Left panel:} Average cut size for
$d$-regular graphs with $d=3$ and $d=5$ as a function of the number of
vertices $n$, bootstrap-averaged over $20$ random graph instances, for
both the GNN-based method and the Goemans-Williamson (GW) algorithm.  On
each graph instance, the GNN solver is allowed up to five shots, and the
GW algorithm takes 100 shots. Solid lines for $n \geq 10^{3}$ represent
theoretical upper bounds, as described in the main text.  Inset: The
estimated relative approximation ratio defined as $\mathrm{cut}^{\star}
/ \mathrm{cut}_{\mathrm{ub}}$ shows that our approach consistently
achieves high-quality solutions.  \textbf{Right panel:} Algorithm
runtime in seconds for both the GNN solver and the GW algorithm. Error
bars refer to twice the bootstrapped standard deviations, sampled across
$20$ random graph instances for every data point.
\label{fig:maxcut}}
\end{figure*}

The complexity of MaxCut depends on the regularity and connectivity of
the underlying graph. Following an existing trend in the community
\citep{guerreschi:19}, we first consider the MaxCut problem on random
(unweighted) $d$-regular graphs, where every vertex is connected to
exactly $d$ other vertices. We perform the benchmarks as follows.  For
graphs with up to a few hundred nodes, we compare our GNN-based solver
to the (approximate) polynomial-time Goemans-Williamson (GW) algorithm
\citep{goemans:95}, which provides the current record for an approximate
answer within some fixed multiplicative factor of the optimum (referred
to as approximation ratio $\alpha$), using semidefinite programming and
randomized rounding.  
Specifically, the GW algorithm achieves a
guaranteed approximation ratio of $\alpha \sim 0.878$ for generic
graphs. This lower bound can be raised for specific graphs such as
unweighted $3$-regular graphs where $\alpha \sim 0.9326$
\citep{halperin:04}. 
Our implementation of the GW algorithm is based on the open-source CVXOPT solver, 
with CVXPY as modeling interface.
For very large graphs with up to a million nodes,
numerical benchmarks are not available, but we can compare our best
solution $\mathrm{cut}^{\star}$ to an analytical result derived in
Ref.~\citep{dembo:17}, where it was shown that with high probability (in
the limit $n \rightarrow \infty$) the size of the maximum cut for random
$d$-regular graphs with $n$ nodes is given by $\mathrm{cut}^{\star} =
(d/4 + P_{*} \sqrt{d/4} + {\mathcal O}(\sqrt{d}))n + {\mathcal O}(n)$.
Here, $P_{*} \approx 0.7632$ refers to an universal constant related to
the ground-state energy of the Sherrington-Kirkpatrick model
\cite{sherrington:75,binder:86} that can be expressed analytically via
Parisi's formula \citep{dembo:17}. We thus take
$\mathrm{cut}_{\mathrm{ub}} = (d/4 + P_{*} \sqrt{d/4})n$ as an
upper-bound estimate for the maximum cut size in the large-$n$ limit.
We complement this upper bound with a lower bound as achieved by a
simple, randomized $0.5$-approximation algorithm that (on average) cuts
half of the edges, yielding a cut size of $\mathrm{cut}_{\mathrm{rnd}}
\approx (d/4) n$ for a $d$-regular graph with $|\mathcal{E}|=(d/2)n$.
Our results for the achieved cut size as a function of the number of
vertices $n$ are shown in Fig.~\ref{fig:maxcut}. All results are
bootstrapped estimates of the mean, with error bars denoting twice the
bootstrapped standard deviations, sampled across 20 random $d$-regular
graphs for every data point. For graphs with up to a few hundred nodes,
we find that a simple two-layer GCN architecture can perform on par with
the GW algorithm, while showing a runtime advantage compared to GW
starting at around $n \approx 100$ nodes. For large graphs with $n
\approx 10^4$ to $10^6$ nodes, we find that our approach consistently
achieves high-quality solutions with $\mathrm{cut}^{\star} \gtrsim 0.9
\cdot \mathrm{cut}_{\mathrm{ub}}$ for both $d=3$ and $d=5$, respectively
(i.e., much better than any naive randomized algorithm).  As expected
for $d$-regular graphs, we find $\mathrm{cut}^{\star}$ to scale linearly
with the number of nodes $n$, i.e., $\mathrm{cut}^{\star} \approx
\gamma_{d} n$, with $\gamma_{3} \approx 1.28$ and $\gamma_{5} \approx
1.93$ for $d=3$ and $d=5$, respectively.  Moreover, utilizing modern GPU
hardware, we observe a favorable runtime scaling at intermediate and
large system sizes that allows us to solve instances with $n = 10^6$
nodes in approximately 10 minutes (which includes both GNN model
training and post-processing steps).  Specifically, as shown in
Fig.~\ref{fig:maxcut}, we observe an approximately linear scaling of
total runtime with $\sim n$, for large $d$-regular graphs with $10^5
\leq n \leq 10^6$; 
contrasted with the observed GW
algorithm scaling as $\sim n^{3.5}$ for problem sizes in the range $n \lesssim 250$,  
thereby showing the (expected) time complexity $\tilde{O}(n^{3.5})$ of the interior-point method (as commonly used for solving the semidefinite program underlying the GW algorithm) 
that dominates the GW algorithm runtime \cite{alizadeh:95, haribara:16}.

\begin{table*}
\centering
 \begin{tabular}{|c | c c | c c c c c | c |} 
 \hline
 graph & nodes & edges & BLS & DSDP & KHLWG & RUN-CSP &  PI-GNN & relative error $\epsilon$ \\ [0.5ex] 
 \hline\hline
 G14 & 800 & 4694 & \textbf{3064} & 2922 & 3061 & 2943 & 3026 &  0.81\%\\ 
 G15 & 800 & 4661 & \textbf{3050} & 2938 & \textbf{3050} & 2928 & 2990 & 1.29\%\\
 G22 & 2000 & 19990 & \textbf{13359} & 12960 & \textbf{13359} & 13028 & 13181 & 0.89\% \\
 G49 & 3000 & 6000 & \textbf{6000} & \textbf{6000} & \textbf{6000} & \textbf{6000} & 5918 & 1.37\%\\
 G50 & 3000 & 6000 & \textbf{5880} & \textbf{5880} & \textbf{5880} & \textbf{5880} & 5820 & 1.00\%\\
 G55 & 5000 & 12468 & \textbf{10294} & 9960 & 10236 & 10116 & 10138 & 1.25\%\\
 G70 & 10000 & 9999 & \textbf{9541} & 9456 & 9458 & --- & 9421 & 1.20\%\\ [1ex] 
 \hline
 \end{tabular}
\caption{
Numerical results for MaxCut on Gset instances.  We report cut sizes
achieved with our physics-inspired GNN solver (PI-GNN), together with
results sourced from Refs.~\cite{toenshoff:19, kochenberger:11,
benlic:13, choi:00}. Best known results are marked in bold.  The last
column specifies the relative error $\epsilon$ comparing PI-GNN to the
best known cut size.  Further details are provided in the main text.
GNN model configurations are detailed in the supplemental material.
\label{tab:gset}} 
\end{table*}

To complement our work on random $d$-regular graphs, we have performed
additional experiments on standard Max-Cut benchmark instances, with
published results, based on the publicly available Gset data set
\cite{ye:03} commonly used for testing Max-Cut algorithms.
We provide benchmark results for seven different graphs,
with thousands of nodes, including (i) two Erd{\"o}s-Renyi graphs with
uniform edge probability, (ii) two graphs where the connectivity
gradually decays from node $1$ to $n$, (iii) two $4$-regular toroidal
graphs, and (iv) one of the largest Gset instances with $n=10^{4}$.  The
results are displayed in Tab.~\ref{tab:gset}.  Here, we report cut sizes
achieved with our physics-inspired GNN solver (PI-GNN), together with
results sourced from Refs.~\cite{toenshoff:19, kochenberger:11,
benlic:13, choi:00}; the latter include an SDP solver using dual scaling
(DSDP) \cite{choi:00}, a combination of local search and adaptive
perturbation referred to as Breakout Local Search (BLS) \cite{benlic:13}
(providing the best known solutions for the Gset data set), a Tabu Search
metaheuristic (KHLWG) \cite{kochenberger:11}, and a recurrent GNN
architecture for maximum constraint satisfaction problems (RUN-CSP)
\cite{toenshoff:19}.  We assess the solution quality achieved with
PI-GNN with the relative error $\epsilon=(\mathrm{cut}_{\mathrm{best}} -
\mathrm{cut}^{\star})/|\mathcal{E}|$ quantifying the gap to the best
known solution, normalized by the number of edges $|\mathcal{E}|$,
thereby giving the fraction of uncut edges as compared to the best known
solution.  We find that our general-purpose approach is competitive with
other solvers and typically within $\sim1\%$ of the best published
results.

\begin{figure*}
\includegraphics[width=2.0 \columnwidth]{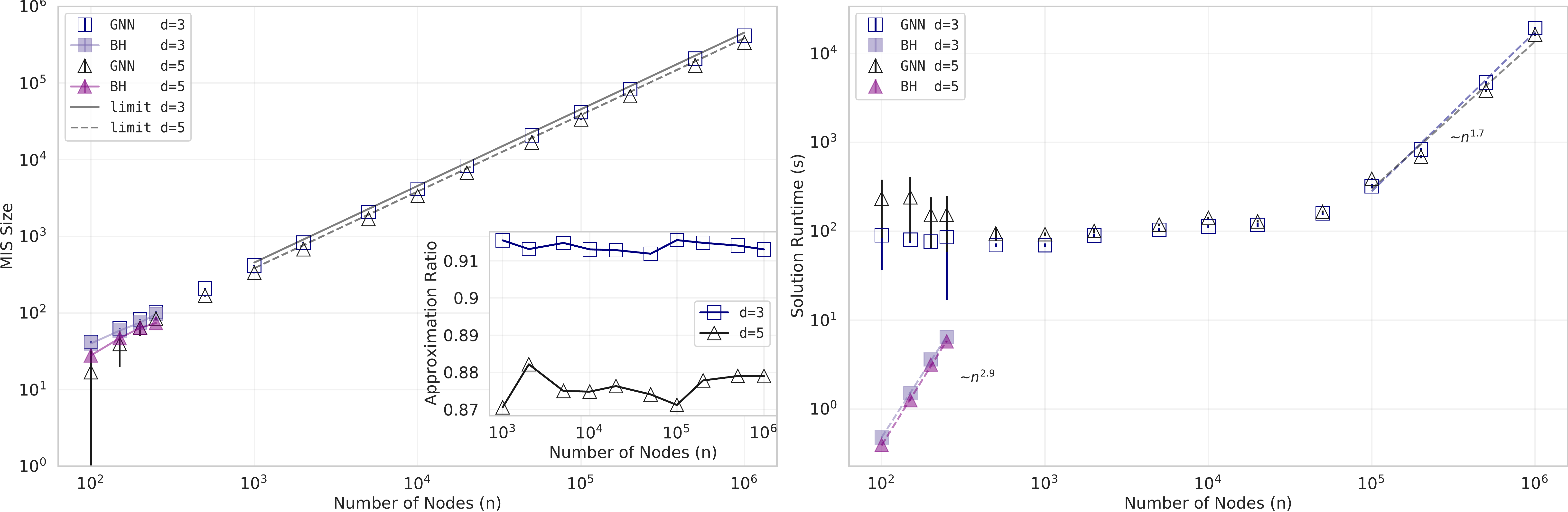}
\caption{
Numerical results for the MIS problem.  \textbf{Left panel}: Average
independence number $\alpha$ for $d$-regular graphs with $d=3$ and $d=5$
as a function of the number of vertices $n$, (bootstrap-)averaged over
$20$ random graph instances, for both the GNN-based method and a
traditional MIS algorithm \citep{boppana:92}.  Solid lines for $n \geq
10^{3}$ refer to theoretical upper bounds as described in the main text.
Inset: The estimated relative approximation ratio comparing the achieved
independence number $\alpha$ against known theoretical upper bounds
shows that our approach consistently achieves high-quality solutions.
\textbf{Right panel}: Algorithm runtime in seconds for both the GNN
solver and the Boppana-Halldorsson algorithm. Error bars refer to twice
the bootstrapped standard deviations, sampled across $20$ random graph
instances for every data point.
\label{fig:mis}}
\end{figure*}

\textbf{Maximum Independent Set.} The MIS problem is a prominent
combinatorial optimization problem with practical applications in
network design \citep{hale:80} and finance \citep{boginski:05}, and is
closely related to the maximum clique, minimum vertex cover, and set
packing problems. In the quantum community, the MIS problem has recently
attracted significant interest \citep{yu:20} as a potential target use
case for novel experimental platforms based on neutral atom arrays
\citep{pichler:18}. The MIS problem reads as follows. Given an
undirected graph $\mathcal{G}=(\mathcal{V}, \mathcal{E})$, an
independent set is a subset of vertices that are not connected with each
other. The MIS problem is then the task to find the largest independent
set, with its (maximum) cardinality typically denoted as the
independence number $\alpha$.  To formulate the MIS problem
mathematically, for a given graph $\mathcal{G}=(\mathcal{V},
\mathcal{E})$, one first associates a binary variable $x_{i} \in
\{0,1\}$ with every vertex $i \in V$, with $x_{i}=1$ if vertex $i$
belongs to the independent set, and $x_{i}=0$ otherwise. The MIS can
then be formulated in terms of a Hamiltonian that counts the number of
marked (colored) vertices and adds a penalty to nonindependent
configurations (when two vertices in this set are connected by an edge).
It is given by
\begin{equation}
H_{\mathrm{MIS}} = -\sum_{i \in \mathcal{V}} x_{i} + P \sum_{(i,j) \in \mathcal{E}} x_{i}x_{j},  \label{eq:MIS}
\end{equation}
with a negative pre-factor to the first term (because we solve for the
largest independent set within a minimization problem), and the penalty
parameter $P>0$ enforcing the constraints. Note that the numerical value
for $P$ is typically set as $P=2$ \citep{djidjev:18}, but can be further
optimized in an outer loop. Energetically, the Hamiltonian
$H_{\mathrm{MIS}}$ favors each variable to be in the state $x_{i}=1$
unless a pair of these are connected by an edge.  Again, the Hamiltonian
$H_{\mathrm{MIS}}$ is quadratic and falls into the broader class of QUBO
problems described by Eq.~\eqref{eq:H-QUBO}; again we provide the
explicit $Q$-matrix for a sample MIS problem in Fig.~\ref{fig:workflow}.

The MIS problem is known to be strongly NP-hard, making the existence of
an efficient algorithm for finding the maximum independent set on
generic graphs unlikely.  In addition, the MIS problem is even hard to
approximate. In general, the MIS problem cannot be approximated to a
constant factor in polynomial time (unless P = NP). Again we study the
MIS problem on random unweighted $d$-regular graphs. Because in our
approach the independence constraint is enforced with soft penalty terms
$\sim P$---just like in any QUBO-based model---the predicted set may
violate the independence condition (i.e., the set may contain nodes
connected by an edge). Setting $P=2$, we have observed these violations
only in very few cases.  If present, as part of our post-processing, we
have enforced the independence constraint by greedily removing one of
the nodes of each induced edge from the set, and only reporting results
after this correction. For small graphs with up to a few hundred nodes,
we compare the GNN-based results to results obtained with the
Boppana-Halldorsson algorithm built into the Python NetworkX
library~\citep{boppana:92}. For very large graphs with up to a million
nodes (where benchmarks are not available) we resort to analytical
upper bounds for random $d$-regular graphs as presented in
Ref.~\citep{duckworth:09}. Here, the best known bounds on the ratio
$\alpha_{d} / n$ are reported as $\alpha_{3} / n = 0.45537$ and
$\alpha_{5} / n = 0.38443$ for $d=3$ and $d=5$, respectively, as derived
using refined versions of Markov's inequality \citep{mckay:87}. Our
results for the achieved independence number as a function of the number
of vertices $n$ are shown in Fig.~\ref{fig:mis}. All results are
bootstrapped estimates of the mean, with error bars denoting twice the
bootstrapped standard deviations, sampled across $20$ random $d$-regular
graphs for every data point.  Our numerical results for MIS are similar
to the observations we have made for MaxCut: for graphs with up to a few
hundred nodes, we find that a simple two-layer GCN architecture can
perform on par with (or better than) the traditional solver, with the
GNN solver showing a favorable runtime scaling. For large graphs with $n
\approx 10^4$ to $10^6$ nodes we find that our approach consistently
achieves high-quality solutions with $\alpha_{3}/n \approx 0.416$ and
$\alpha_{5}/n \approx 0.338$ for $d=3$ and $d=5$, respectively,
resulting in estimated numerical approximation ratios of $0.416/0.45537
\sim 0.92$ and $0.338/0.38443 \sim 0.88$, respectively.  Finally, as
shown in Fig.~\ref{fig:mis}, we observe a moderate, super-linear scaling
of the total runtime as $\sim n^{1.7}$ for large $d$-regular graphs with
$n \gtrsim 10^5$, as opposed to the Boppana-Halldorsson solver with a
runtime scaling of $\sim n^{2.9}$ in the range $n \lesssim 500$.  Note
that the GNN model training alone displays sub-linear runtime scaling as
$\sim n^{0.8}$, in line with our MaxCut results, while the aggregate
runtime (including post-processing to enforce the independence
condition) scales as $\sim n^{1.7}$ in the regime $n \sim 10^5 - 10^6$.

\section{Applications in Industry}
\label{Applications}

While our previous analysis has focused on canonical graph optimization
problems such as maximum cut and maximum independent set, in this
section we discuss real-world applications in industry for which our
solver could provide solutions, in particular at potentially
unprecedented problem scales. We focus on applications of the QUBO
formalism, even though our methodology is not limited to this modeling
framework. We first review the existing literature, providing relevant
references across a wide stack of problem domains.  Thereafter, we
explicitly show how to distill combinatorial QUBO problems for a few
select real-world use cases, from risk diversification in finance to
sensor placement problems in water distribution networks. Once in QUBO
format, the problem can be plugged into our general-purpose
physics-inspired GNN solver, as outlined above.

As extensively reviewed in Refs.~\citep{glover:18},
\cite{kochenberger:14}, and \cite{lucas:14}, the QUBO (or, equivalently,
Ising) formalism provides a comprehensive modeling framework
encompassing a vast array of optimization problems, including knapsack
problems, task (resource) allocation problems, and capital budgeting
problems, among others.  Specifically, the applicability of the QUBO
representation has been reported for problem settings involving circuit
board layouts \citep{groetschel:88}, capital budgeting in financial
analysis \citep{laughhunn:70}, computer aided design (CAD)
\citep{krarup:78}, electronic traffic management \citep{gallo:80,
witsgall:75}, cellular radio channel allocation \citep{chardaire:94},
molecular conformation \citep{phillips:94}, and the prediction of
epileptic seizures \citep{iasemidis:03}, among others. As mentioned
earlier, practical applications of the MaxCut problem can be found in
machine scheduling \citep{alidaee:94}, image recognition
\citep{neven:08}, and electronic circuit layout design \citep{deza:94}.
Similarly, in what follows we discuss in detail three select use cases
and how they can be cast in QUBO form and thus made amenable to our
solver.

\textbf{Risk diversification.} Graphs offer a convenient framework to
model portfolio management problems in finance.  Specifically, here we
outline a risk diversification strategy, but similar considerations
apply for the implementation of hedging strategies \citep{kalra:08}.  We
consider a (potentially very large) universe of $n$ assets, for which we
are given a vector $\boldsymbol{\mu} \in \mathbb{R}^{n}$ describing
expected future returns, and the covariance matrix $\Sigma \in
\mathbb{R}^{n \times n}$ capturing volatility through the correlations
among assets. To minimize the volatility of returns of our portfolio,
our goal is to select a subset of uncorrelated assets with the largest
possible diversified portfolio.  To this end we consider a graph
$\mathcal{G}$ with $n$ nodes, with every node representing one asset.
Correlations can be described in graph form, either by directly taking
the cross-correlation matrix as a weighted adjacency matrix, or by
creating a binary adjacency matrix $A$ through thresholding. We set
$A_{i,j}=1$ if and only if the absolute value of the correlation between
assets $i$ and $j$ is greater than some user-specific threshold
parameter $\lambda$, and $A_{i,j}=0$ otherwise \citep{kalra:08}.
Accordingly, in our model pairs of assets are classified as correlated
or uncorrelated, based on whether or not the corresponding correlation
coefficient exceeds a minimum level.  Overall, the risk diversification
strategy outlined above can be cast as an optimization problem in QUBO
form, with the Hamiltonian
\begin{equation}
H_{\mathrm{wMIS}} = - \sum_{i \in \mathcal{V}} \mu_{i} x_{i} + 
P \sum_{(i,j) \in \mathcal{E}} x_{i}x_{j}, 
\end{equation}
in a straightforward extension of Eq.~\eqref{eq:MIS} from the standard
MIS problem to the weighted MIS problem.  Accordingly, by minimizing the
first term of $H_{\mathrm{wMIS}}$ we pick large-return assets subject to
the independent set constraint (as captured by the second term).  This
diversification model is reminiscent of the mean-variance Markowitz
model \citep{markowitz:52}, albeit in discretized form.  Given our
results for the standard MIS problem with small tweaks to the loss
function, such a problem could be readily plugged into our solver, for
example as part of a larger, two-stage portfolio management pipeline
where first a subset of assets is selected from a larger universe of
assets using our solver, and then capital is allocated within a smaller,
sparsified basket of assets using off-the-shelf solvers.

\textbf{Interval scheduling.} Here we consider a scenario involving the
scheduling of tasks with given start and end times, as relevant for
example in the context of algorithm design in computer science
\citep{kolen:07}.  Specifically, we face $n$ resource requests, each
represented by an interval specifying the time in which it needs to be
processed by some machine. Typically some requests will overlap in time
leading to request clashes that cannot be satisfied by the same machine
(resource).  Conversely, a subset of intervals is deemed compatible if
no two intervals overlap on the machine.  As commonly done in resource
allocation problems and scheduling theory, this situation can
conveniently be described with the help of an undirected
\textit{interval graph} $\mathcal{G}$ in which we introduce a vertex for
each request and edges between vertices whose requests overlap.  With
the goal to maximize the throughput (i.e., to execute as many tasks as
possible on a single machine), the interval scheduling maximization
problem is then to find the largest compatible set, that is, a set of
non-overlapping intervals of maximum size.  This use case is equivalent
to finding the maximum independent set in the corresponding interval
graph $\mathcal{G}$ with $n$ nodes.  While inexpensive (special-purpose)
algorithms exist for interval graphs \citep{bar-noy:01}, we can then
solve the underlying MIS problem, as described by Eq.~\eqref{eq:MIS}, on
this interval graph, in the same way as any other QUBO problem, using
our general-purpose GNN-based approach, following the methodology
outlined above.

\textbf{Sensor placement in water distribution networks.} Optimal sensor
placement is key to the detection and isolation of fault events---such
as water leaks---in water distribution networks (WDN)
\citep{speziali:21}. As detailed in Ref.~\citep{speziali:21}, the
problem of optimally placing pressure sensors on a WDN can be
efficiently cast as QUBO problem. Specifically, a WDN can be readily
mapped to a graph $\mathcal{G}=(\mathcal{V}, \mathcal{E})$, with the
nodes $i \in \mathcal{V}$ referring to tanks or junctions and edges
$(i,j) \in \mathcal{E}$ representing pipes, respectively.  We then
associate a binary variable $x_{i}=0,1$ with every node, and set
$x_{i}=1$ if node $i$ hosts a sensor, and $x_{i}=0$ otherwise.  The
problem of covering the WDN with the smallest possible number of
pressure sensors then maps onto the minimum vertex cover problem
\citep{speziali:21}, as described by the Hamiltonian
\begin{equation}
H_{\mathrm{MVC}} = \sum_{i \in \mathcal{V}} c_{i} x_{i} + P
\sum_{(i,j) \in \mathcal{E}} (1-x_{i} - x_{j} +x_{i} x_{j}).
\end{equation}
Here, $c_{i} \geq 0$ denotes the cost of node $i$ hosting a sensor, the
first term describes the overall cost of any potential sensor placement
strategy, while the second (penalty) term with $P>0$ ensures the
constraint $x_{i} + x_{j} \geq 1$ for all edges $(i,j) \in \mathcal{E}$
(i.e., at least one of the endpoints of each edge will be in the cover
\citep{glover:18}). Potential tweaks to this model are detailed in
Ref.~\citep{speziali:21}, however, variations can all be represented as
a QUBO problem. Along the lines of our previous analysis for the MaxCut
or MIS problem, the Hamiltonian $H_{\mathrm{MVC}}$ (or some variation
thereof) can be straightforwardly mapped to a relaxed loss function with
which we can train our solver and then solve the corresponding sensor
placement use case.

\section{Conclusion and Outlook}
\label{Conclusion}

In summary, we have proposed and analyzed a versatile and scalable
general-purpose solver that is powered by graph neural networks and
draws from concepts in statistical physics. Our approach is applicable
to any $k$-local Ising model, including canonical NP-hard combinatorial
optimization problems such as the maximum cut, maximum clique, minimum
vertex cover or maximum independent set problems, among others
\citep{lucas:14}. Starting from a problem formulation in Ising form, we
apply a relaxation strategy to the problem Hamiltonian by dropping
integrality constraints on the decision variables in order to generate a
differentiable loss function with which we perform unsupervised training
on the node representations of the GNN. The GNN is then trained to
generate soft assignments to predict the likelihood of belonging in one
of two classes, for each vertex in the graph. To find a binary
(two-color) labelling consistent with the original problem formulation,
simple projection heuristics are applied. Overall, we find that this
approach can compete with existing special-purpose solvers, such as the
Goemans-Williamson algorithm designed to solve the maximum cut problem,
with the potential to tap into the rich toolbox of statistical physics,
including, for example, the study of phase transitions. In the current
noisy intermediate scale quantum era, our approach could be used as a
broadly applicable, scalable benchmark for emerging quantum
technologies, including special-purpose quantum \cite{johnson:11} and
quantum-inspired annealers \cite{matsubara:17}, while not being resource
constrained nor being limited to problem instances in QUBO form, as is
also the case for coherent Ising machines \citep{mahon:16}.

Finally, we highlight possible extensions of research going beyond our
present work. First, to better understand the limitations of GNNs in the
context of combinatorial optimization, further studies are in order,
systematically benchmarking GNNs against state-of-the-art solvers for a
large class of optimization problems while leveraging the entire zoo of
GNN implementations including, for example, GraphSAGE
\citep{hamilton:17} or Graph Attention Networks (GATs)
\citep{velickovic:18} to potentially boost the GNN Ansatz with an
attention mechanism enabling vertices to weigh neighbor representations
during the aggregation steps.  
Second, the presented GNN approach should
be able to accommodate problems sizes with hundreds of millions of nodes
when leveraging distributed training in a mini-batch fashion on a
cluster of machines \citep{zheng:20}, thereby challenging the
capabilities of several existing solvers.  
While we have solved
individual problem instances from scratch, using a random initialization
process for the initial node embeddings, in the future warm-starting the
training process with pre-trained weights (transfer learning) could
boost the time to solution. 
Moreover, one could potentially boost the performance of our optimizer by implementing randomized projection schemes (as opposed to the simple deterministic approach used here), 
or augment these strategies with simple greedy post-processing routines that check for local optimality with a sequence of local bit flips. 
Finally, as discussed in the main text, our
approach can be generalized to PUBO problems on hyper-graphs where
so-called hyper-edges may contain more than just two nodes, with no need
for (typically) resource-intensive degree reduction schemes, as opposed
to resource-constrained QUBO solvers.  Potential applications cover many
real-world optimization problems involving multi-body interactions, as
found in scheduling problems \citep{bansal:10} or chemistry
\citep{hernandez:16, terry:19}.  In conclusion, the proposed
cross-fertilization between machine learning, operations research and
physics opens up a number of interesting research directions, with the
ultimate goal to further advance our ability to solve hard combinatorial
optimization problems.

\subsection*{Data availability}

The data necessary to reproduce our numerical benchmark results are
publicly available at https://web.stanford.edu/$\sim$yyye/yyye/Gset/. Random
d-regular graphs have been generated using the open-source networkx
library (https://networkx.org).

\subsection*{Code availability}

Code availability statement: An end-to-end open source demo version of
the code implementing our approach has been made publicly available at
https://github.com/amazon-research/co-with-gnns-example.

\bibliography{refs}

\begin{thebibliography}{117}
\bibitem[{cod()}]{code}
\bibinfo{note}{A working code example can be obtained at
  {{{github.com/amazon-research/co-with-gnns-example}}}}.
\end{thebibliography}

\noindent {\bf Acknowledgments} We thank Fernando Brandao, George
Karypis, Michael Kastoryano, Eric Kessler, Tyler Mullenbach, Nicola
Pancotti, Mauricio Resende, Shantu Roy, Grant Salton, Simone Severini,
Ayinger Urweisse, and Jason Zhu for fruitful discussions.

\vspace{1em}

\noindent {\bf Author contributions} All authors contributed to the
ideation and design of the research.  M.J.A.S.~and J.K.B.~developed and
ran the computational experiments, as well as wrote the initial draft of
the the manuscript. H.G.K.~supervised this work and revised the
manuscript.

\vspace{1em}

\noindent {\bf Competing interests} M.J.A.S., J.K.B.~and H.G.K.~are
listed as inventors on a US provisional patent application (number
7924-38500) on combinatorial optimization ith graph neural networks.

\vspace{1em}

\noindent {\bf Additional information} Correspondence and requests for
materials should be addressed to M.J.A.S., J.K.B. or H.G.K.


\newpage \onecolumngrid \newpage { \center \bf \large  Supplemental Material for: \\  Combinatorial Optimization with Physics-Inspired Graph Neural Networks \vspace*{0.1cm}\\  \vspace*{0.0cm} } 
\begin{center} Martin J. A. Schuetz,$^{1,2,3}$ J. Kyle Brubaker,$^{2}$  and Helmut G. Katzgraber$^{1,2,3}$ \\ 
\vspace*{0.15cm} 
\small{\textit{$^{1}$Amazon Quantum Solutions Lab, Seattle, Washington 98170, USA}} \\
\small{\textit{$^{2}$AWS Intelligent and Advanced Compute Technologies, Professional Services, Seattle, Washington 98170, USA}} \\
\small{\textit{$^{3}$AWS Center for Quantum Computing, Pasadena, CA 91125, USA}} 
\vspace*{0.25cm} 
\end{center}


\setcounter{section}{0}

\setcounter{equation}{0} 
\setcounter{figure}{0} 
\setcounter{page}{1} 
\makeatletter 
\renewcommand{\theequation}{S\arabic{equation}} 
\renewcommand{\thefigure}{S\arabic{figure}} 
\renewcommand{\bibnumfmt}[1]{[S#1]} 
\renewcommand{\citenumfont}[1]{S#1} 


\section{Core GCN code block} \label{code-block}

\begin{mdframed}[backgroundcolor=gray!10]
\lstset{language=Python}
\lstset{frame=lines}
\lstset{caption={Core code block of example script based on the DGL
library. The first block defines a two-layer GCN architecture Ansatz;
the second code block defines the loss function as described by
Eq.~\eqref{eq:loss-QUBO}. Further details can be found in the main text,
as well as in Ref.~\cite{code}.}}
\lstset{label={listing:code-block}}
\lstset{basicstyle=\footnotesize}
\begin{lstlisting}
# Import required packages
import dgl
import torch
import torch.nn as nn
from dgl.nn.pytorch import GraphConv

# Define two-layer GCN
class GCN(nn.Module):
  def __init__(self,in_feats,hidden,classes):
      super(GCN,self).__init__()
      self.conv1 = GraphConv(in_feats,hidden)
      self.conv2 = GraphConv(hidden,classes)

  def forward(self,g,inputs):
      h = self.conv1(g,inputs)
      h = torch.relu(h)
      h = self.conv2(g,h)
      # binary classification
      h = torch.sigmoid(h)  
      return h

# Define custom loss function for QUBOs
def loss_func(probs_,Q_mat):
  """
  function to compute cost value for given 
  soft assignments and predefined QUBO matrix
  """
  # minimize cost = x.T * Q * x
  cost = (probs_.T @ Q_mat @ probs_).squeeze()
    
  return cost
\end{lstlisting}
\end{mdframed}

\newpage

\section{Hyperparameters for G-Set Experiments} \label{Gset-hypers}

In this section, we provide details for the specific model
configurations (hyperparameters) as used to solve the Gset instances
with our physics-inspired GNN solver (PI-GNN).  The results achieved
with these model configurations are displayed in Tab.~\ref{tab:gset};
the corresponding hyperparameters are given in
Tab.~\ref{tab:gset-hypers}.  Our base GCN architecture with tunable
number of layers $K$ is specified in Listing \ref{listing:code-GCN}.

\begin{table*}[h]
\centering
 \begin{tabular}{|c | c  | c c c c c c c |} 
 \hline
 graph & PI-GNN & embedding $d_{0}$ & layers $K$ & hidden dim $d_1$ & hidden dim $d_2$ & hidden dim $d_3$ & learning rate $\beta$ & dropout \\ [0.5ex] 
 \hline\hline
 G14 & 3026 & 369 & 1 & 5 & --- & --- & 0.00467 & 0.0 \\ 
 G15 & 2990 & 394 & 1 & 5 & --- & --- & 0.00587 & 0.0 \\
 G22 & 13181 & 419 & 2 & 1909 & 3401 & --- & 0.00103 & 0.4498 \\
 G49 & 5918 & 2167 & 3 & 2338 & 1955 & 8 & 0.00058 & 0.3554 \\
 G50 & 5820 & 208 & 3 & 218 & 3582 & 566 & 0.00488 & 0.2365 \\
 G55 & 10138 & 278 & 3 & 8412 & 8352 & 5499 & 0.00161 & 0.1062 \\
 G70 & 9421 & 109 & 3 & 1233 & 7048 & 11869 & 0.00139 & 0.3912 \\ [1ex] 
 \hline
 \end{tabular}
\caption{
Numerical results for MaxCut on Gset instances, with hyperparameters specified for the PI-GNN solver.
\label{tab:gset-hypers}}
\end{table*}

\begin{mdframed}[backgroundcolor=gray!10]
\lstset{language=Python}
\lstset{frame=lines}
\lstset{caption={Base GCN architecture used for solving MaxCut on Gset problem instances.}}
\lstset{label={listing:code-GCN}}
\lstset{basicstyle=\footnotesize}
\begin{lstlisting}
# Define GNN object
class GCN_dev(nn.Module):
    def __init__(self, in_feats, hidden_sizes, dropout, num_classes):
        super(GCN_dev, self).__init__()
        # Combine all layer sizes into a single list
        all_layers = [in_feats] + hidden_sizes + [num_classes]  
        # slice list into sub-lists of length 2
        self.layer_sizes = list(window(all_layers)) 
        # reference to ID final layer
        self.out_layer_id = len(self.layer_sizes) - 1  
        self.dropout_frac = dropout
        self.layers = OrderedDict()
        for idx, (layer_in, layer_out) in enumerate(self.layer_sizes):
            self.layers[idx] = GraphConv(layer_in, layer_out).to(DEVICE)
    def forward(self, g, inputs):
        for k, layer in self.layers.items():
            if k == 0: # reference to ID final layer
                h = layer(g, inputs)
                h = torch.relu(h)
                h = F.dropout(h, p=self.dropout_frac)
            elif 0 < k < self.out_layer_id: # intermediate layers
                h = layer(g, h)
                h = torch.relu(h)
                h = F.dropout(h, p=self.dropout_frac)
            else: # output layer
                h = layer(g, h)
                h = torch.sigmoid(h)  # binary classification
        return h

\end{lstlisting}
\end{mdframed} 

\end{document}